\documentclass[prx, reprint, amsmath, amssymb, longbibliography, superscriptaddress, floatfix, nofootinbib]{revtex4-2}

\usepackage[utf8]{inputenc}
\usepackage{graphicx}

\usepackage[svgnames]{xcolor}
\usepackage{amsmath,amssymb,mathrsfs}
\usepackage{bm}
\usepackage[unicode=true, pdfusetitle, bookmarks=true, bookmarksnumbered=false, bookmarksopen=false, breaklinks=false, pdfborder={0 0 1}, backref=false, colorlinks=false]
 {hyperref}
\usepackage{caption}
\usepackage{subcaption}

\usepackage{ragged2e}
\DeclareCaptionJustification{justified}{\justifying}
\usepackage{diagbox}
\usepackage{braket,tensor,siunitx}
\usepackage[inkscapeformat=png]{svg}
\usepackage{soul}
\usepackage{cancel}

\usepackage[percent]{overpic}
\usepackage{tikz}
\usepackage{array}
\usepackage{multirow}

\newcommand*\showwidth[1]{%
  \textcolor{blue}{\rule{\csname#1\endcsname}{1pt}}\newline
  \texttt{\textbackslash#1}: \expandafter\the\csname#1\endcsname
  \par
}

\makeatletter
\newcommand\thefontsize{The current font size is: \f@size pt}
\makeatother

\begin{document}
 
\title{Fisher Information Flow in Artificial Neural Networks}
\author{Maximilian Weimar}
\email{maximilian.weimar@tuwien.ac.at}
\affiliation{Institute for Theoretical Physics, Vienna University of Technology (TU Wien), A–1040 Vienna, Austria}
\author{Lukas M. Rachbauer}
\affiliation{Institute for Theoretical Physics, Vienna University of Technology (TU Wien), A–1040 Vienna, Austria}
\author{Ilya Starshynov}
\affiliation{School of Physics and Astronomy, University of Glasgow, Glasgow, G12 8QQ, United Kingdom}
\author{Daniele Faccio}
\affiliation{School of Physics and Astronomy, University of Glasgow, Glasgow, G12 8QQ, United Kingdom}
\author{Linara Adilova}
\affiliation{Faculty of Computer Science, Ruhr University Bochum, 44801 Bochum, Germany}
\author{Dorian Bouchet}
\affiliation{Universit{\'e} Grenoble Alpes, CNRS, LIPhy, 38000 Grenoble, France}
\author{Stefan Rotter}
\affiliation{Institute for Theoretical Physics, Vienna University of Technology (TU Wien), A–1040 Vienna, Austria}
\date{\today}

\begin{abstract}
The estimation of continuous parameters from measured data plays a central role in many fields of physics. A key tool in understanding and improving such estimation processes is the concept of Fisher information, which quantifies how information about unknown parameters propagates through a physical system and determines the ultimate limits of precision. With artificial neural networks (ANNs) gradually becoming an integral part of many measurement systems, it is essential to understand how they process and transmit parameter-relevant information internally.  Here, we present a method to monitor the flow of Fisher information through an artificial neural network performing a parameter estimation task, tracking it from the input to the output layer. We show that optimal estimation performance corresponds to the maximal transmission of Fisher information, and that training beyond this point results in information loss due to overfitting. This provides a model-free stopping criterion for network training-eliminating the need for a separate validation dataset. To demonstrate the practical relevance of our approach, we apply it to a network trained on data from an imaging experiment, highlighting its effectiveness in a realistic physical setting.
\end{abstract}

\maketitle

\section{Introduction}
\label{Sec:Introduction}
In physics, artificial neural networks (ANNs) are used in a wide variety of research areas, ranging from optics \cite{yoon2020deep} to high-energy physics \cite{shlomi2020graph}, the computation of material parameters \cite{xie2018crystal}, the design of experiments \cite{Krenn2016}, and the solution of complex inverse problems \cite{jin2017deep}. In the context of ANN training, one of the central issues is to understand how models process the information that they receive as input~\cite{cover1999elements,mackay2003information,tishby2000information,geiger2020information}.

One particularly successful approach is based on the notion of mutual information (MI) and the information bottleneck method~\cite{IB_DL,IB_overview,geiger2021information}, which provide a peek into the training process of the model. MI is also used for regularizing the optimization~\cite{kolchinsky2019nonlinear, wu2020learnability} and for deriving generalization bounds~\cite{neu2021information, harutyunyan2021information}. Nevertheless, this line of research is severely hindered by the challenges of estimating MI in high-dimensional spaces~\cite{mcallester2020formal} and by the issue that the MI diverges for continuous random variables when they are connected by a deterministic transformation \cite{saxe2019information, MI_Flow, lorenzen2021information, adilova2022information}.

The starting point for the current work is the observation that a whole class of estimation problems in physics centers around the concept of Fisher information (FI) \cite{parameter_estimation_literature,properties}. Arising from the field of (statistical) estimation theory \cite{parameter_estimation_literature}, FI is the key quantity when dealing with estimating continuous parameters from noisy data. The amount of FI one has available on a given parameter ultimately determines how precisely the value of this parameter can be estimated and, thus, bounds the achievable performance of the estimating model. As FI applies by definition to the estimation of continuous deterministic variables it naturally avoids the divergence that occurs for the MI in deterministic ANNs. Parameter estimation tasks of this kind are particularly relevant in optics and photonics, such as in imaging \cite{barrett2004foundations,optimal_point_spread_for_imaging,deep_speckle_correlations,dynamic_light_scattering}, particle levitation \cite{gonzalez2021levitodynamics,hupfl2024continuity}, and classical and quantum metrology \cite{QFI,learning_quantum_systems,LIGO,Jezek2003}, to name just a few. 

With the increased use of ANNs to solve these parameter estimation problems, it is, thus, of crucial importance to understand how FI (rather than MI) behaves as the data are being processed by the network. Insights in this direction will help us to understand the limitations of ANNs in solving estimation tasks; they can also be expected to provide an essential tool for the design of new network architectures, which perform better in terms of the preservation of information during data processing. Consider as a specific example that in many scenarios, where a parameter is estimated from high-dimensional image data, the data dimensionality needs to be strongly reduced. Analyzing the FI may reveal how an ANN solves the nontrivial task of preserving information during this compression process.

Despite the relevance of FI in ANNs, the analysis of the FI flow through the network's layers has, to the best of our knowledge, not been provided yet. Previous works \cite{natural_gradient,FI_wrt_parameters1,FI_wrt_parameters2,EWC,achille2019critical} have focused on the FI of an ANN's parameters such as its weights and biases, which is conceptually different from the FI about parameters inherent to the data itself. Moreover, these earlier works invoke assumptions about the structure of the ANNs or the shape of the distributions. In practice, however, modeling the distributions of the data that are fed into an ANN is very difficult, as the network can generate high-dimensional and arbitrarily complicated representations of the data \cite{lecun2015deep,python_DL}. The dimensionality of the data inside a layer of the network is equal to the number of the layer's nodes, which is often of the order of several hundreds even in comparably small models. The FI, however, is a functional of the probability density function describing the data distribution, and there is no feasible way of estimating such a high-dimensional density when it does not take on a very simple form (such as a multivariate Gaussian). In other words, evaluating how much FI is contained in each layer of the ANN is a challenging problem.  
\begin{figure*}[th!]
\centering
\begin{overpic}[width=0.8\textwidth]{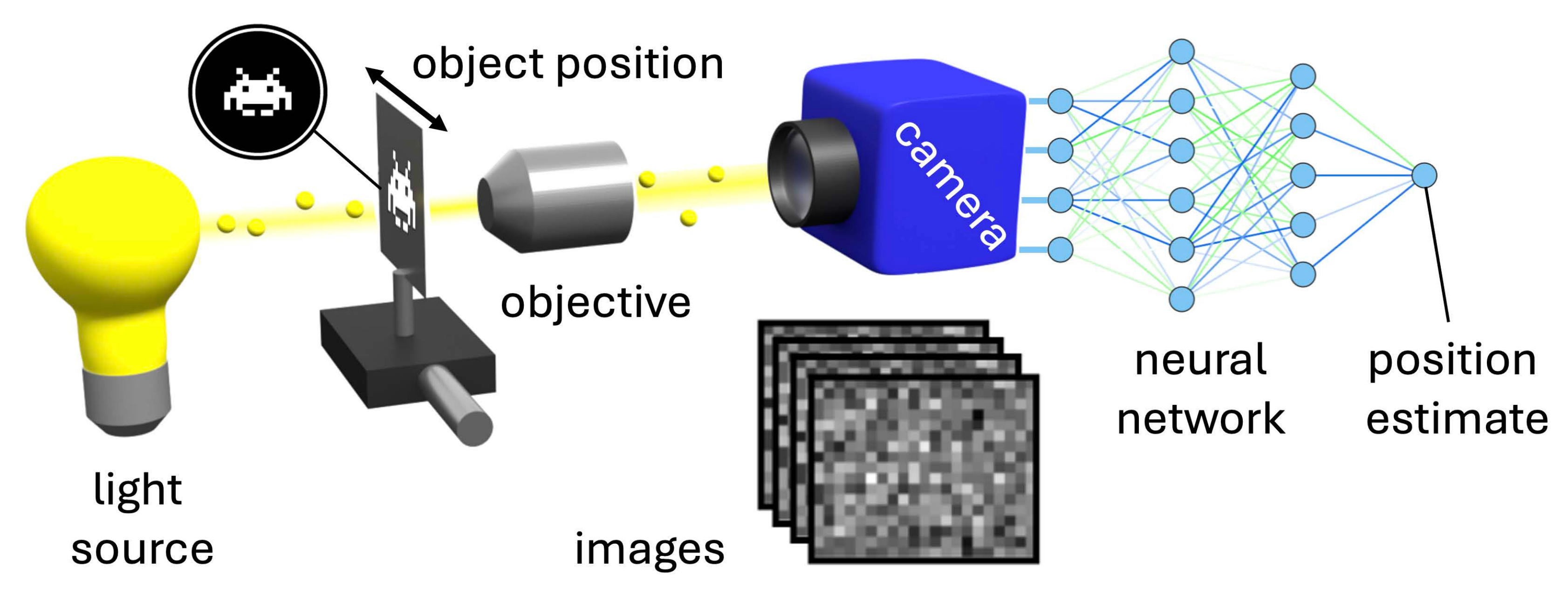}
\end{overpic}
 \caption{A beam of light, originating from the source on the left, is directed onto the target object – the image of a ``space invader'' placed on a slide orthogonal to the beam. The intensity of the light field transmitted through an objective to a camera carries FI about the parameter of interest $\theta$, which is here the horizontal position of the target object. The image data collected in the experiment are fed into an ANN, which is trained to make a prediction $\hat{\theta}$ of the parameter $\theta$ implicitly contained in the data. For the neural network to achieve satisfactory estimates of $\theta$, the FI must be preserved as much as possible while passing through the ANN from input to output layer.}
 \label{fig:FI_Flow}
\end{figure*}

Here, we solve this problem by using the concept of ``linear Fisher information'' \cite{LFI_1984,LOLE} (LFI) for estimating the true FI directly from the dataset – on all of the network's layers and for arbitrarily complicated distributions. To demonstrate the practical relevance of this approach, we study an ANN trained on data from an optical imaging experiment performed under strong noise conditions. At different stages of the training, we observe how the FI behaves as the data are processed by the network. Using the LFI analysis, we are able to demonstrate how FI flows through an ANN from input to output. When being well trained, the ANN is able to preserve FI even when considerable dimensionality reduction takes place. At this point the ANN performs near the ultimate limit of precision that is determined by the physical noise within the measured data. Based on this insight, we are able to show that by calculating the FI contained in the ANN's input data, we can determine – using only the training data and no validation data – when the network starts to overfit in the training process.

\section{PHYSICAL SYSTEM AND FISHER INFORMATION ANALYSIS}
\label{Sec:PHYSICAL SYSTEM AND FISHER INFORMATION ANALYSIS}

\subsection{Imaging experiment}
\label{subsec:Imaging experiment}

To perform the proposed FI analysis with a practical parameter estimation task, we consider an imaging experiment subject to strong noise, see Fig.~\ref{fig:FI_Flow}. The data collected during the experiment are used to train an ANN, which we analyze using FI.

\begin{figure}[th!]
\centering
\includegraphics[width=0.5\textwidth]{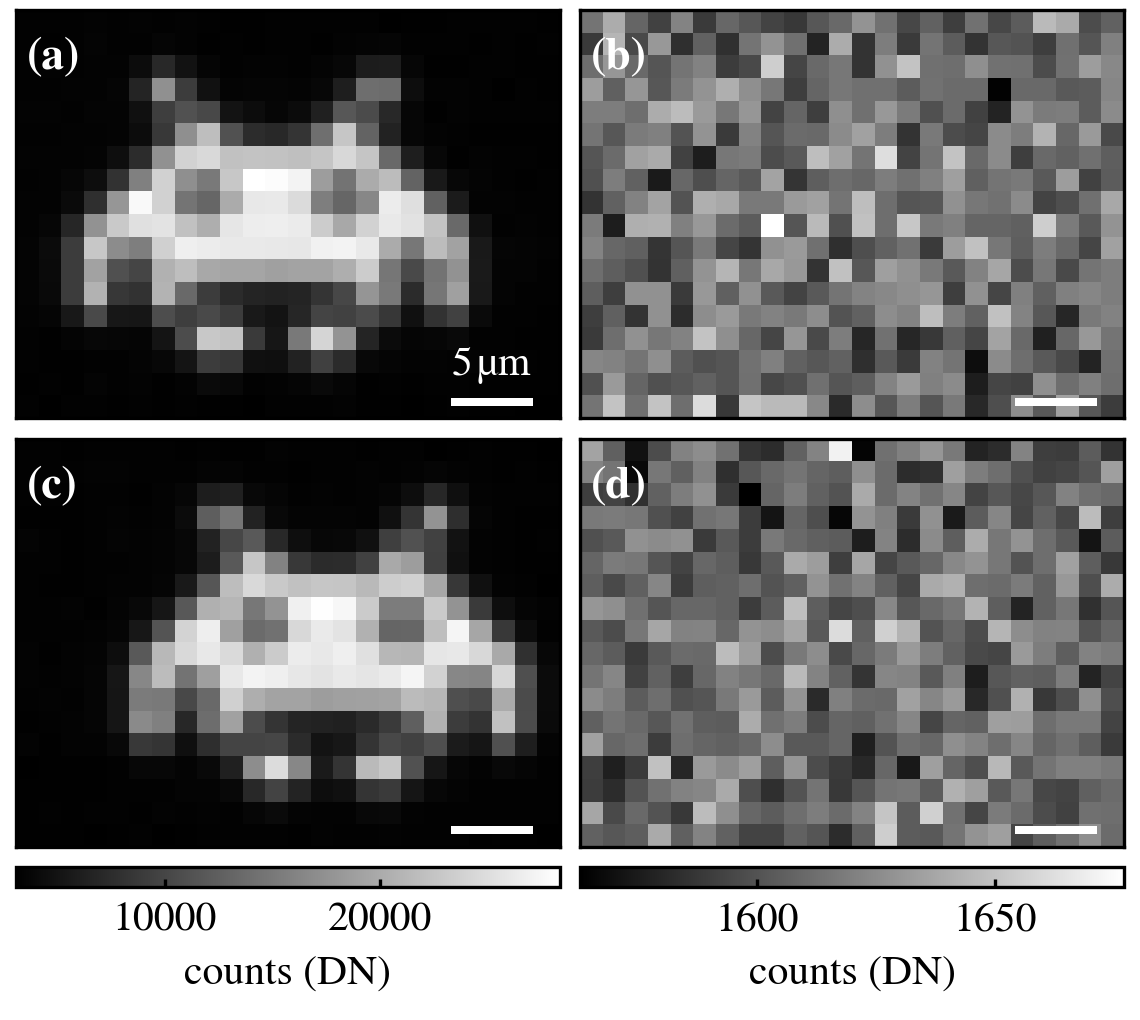}
 \caption{Camera images of a ``space invader'' illuminated by broadband light. The position of the object under study is adjusted by moving the slide to which it is attached (in the horizontal direction). The leftmost position is shown in the top row (a),(b) and the rightmost position in the bottom row (c),(d). The left column (a),(c) shows the camera image under bright illumination and the right column (b),(d) under weak illumination conditions. The digital numbers (DN) that the camera outputs as photons are impinging are shown in the grayscale bars at the bottom.}
 \label{fig:spaceInvader}
\end{figure}

The parameter of interest is the horizontal position of an object (``space invader'' etched in an opaque slide), that can be freely translated in the plane of the slide. We illuminate it with broadband light and collect a series of images at different positions of the object, which we then use to train the ANN to estimate this position (see Appendix \ref{sec:experimental_methods} for details about the experimental setup). We attenuate the incident light up to a level when the camera noise becomes dominant over the image signal, as shown in Fig.~\ref{fig:spaceInvader} (resulting SNR of around $13\%$), which leads to a sizable uncertainty of the position estimates. The statistical distribution of the intensity values measured by each camera pixel is nontrivial (see Appendix \ref{subsec:Sources of noise}), preventing us from constructing an efficient estimator of the position using an analytical expression. The ultimate precision at which this position can be estimated is given by the FI of the input data probability distribution. We calculate the FI of the hidden layers of the network and demonstrate that in the process of training, it gains the ability to transfer FI from the input to the output, thereby achieving this limit.

\subsection{Fisher information and the Cram\'er-Rao lower bound}
\label{subsec:Fisher Information and CRLB}

Let us first recall some insights from estimation theory \cite{parameter_estimation_literature,van2004detection,parameter_estimation}: Consider a parametric family of distributions $p(X;\theta)$ with $X$ being a random variable of arbitrary dimension, which would represent a measured image in the experiment that we described in the previous section. The shape and the spread of the image's distribution are governed by the noise. The deterministic parameter $\theta$ that this distribution depends on is in our case the horizontal position of the object. The FI \cite{parameter_estimation_literature,properties} contained in this distribution is defined as
\begin{equation}
I(\theta) = \langle  \left\{ \partial_{\theta} \mathrm{log} [p(X;\theta)]\right\}^2  \rangle \;,
    \label{eqn:FI}
\end{equation}
where the average $\langle\cdots\rangle$ is taken with respect to the data statistics. A function $\hat{\theta} = \tau(X)$ that assigns a prediction $\hat{\theta}$ of the parameter $\theta$ to the data is referred to as an estimator \cite{parameter_estimation_literature}. Restricting ourselves here to the class of unbiased estimators (where $\langle \hat{\theta} -\theta \rangle = 0$ holds), the FI bounds the achievable precision of the estimate \cite{parameter_estimation,properties} by 
\begin{equation}
    \sigma^2 \geq \frac{1}{I}\;,
    \label{eqn:CRLB}
\end{equation}
with $\sigma^2$ being the variance of $\hat{\theta}$. While this Cram\'er-Rao lower bound (CRLB) can be generalized also to multiple parameters \cite{van2004detection}, we focus here only on the single parameter case, for the sake of simplicity.

\subsection{Fisher information inside deep neural networks}
\label{subsec:Fisher Information within Deep Neural Networks}
The ANN we use for the parameter estimation task can be seen as a nonlinear function $\tau(X)$ combining multiple functions from the layers of the network \cite{lecun2015deep,python_DL}. We feed a set of data to the network and extract the transformed data at the consecutive layers. In a feedforward neural network architecture without skip connections, the transformed data in the hidden layers form a Markov chain \cite{FII,IB_DL,IB_DL_2}. The so-called Fisher information inequality \cite{FII}, a special case of the data processing inequality, assures that the FI does not increase in a Markov chain, i.e., it is a nonincreasing function of the ANN's layer depth \cite{FII,properties}. The causes for the loss of FI in ANNs are discussed in Appendix \ref{subsec:Sufficient Statistics and Efficiency}. 

Evaluating Eq.~(\ref{eqn:FI}) by inferring the probability density function $p(X;\theta)$ is not feasible, as the random variables we encounter in the imaging experiment are high dimensional and conventional methods of probability density estimation are hampered by the curse of dimensionality; i.e., the complexity of the problem grows rapidly with increasing dimension. To circumvent this problem, we use the LFI \cite{TUM_LFI,CF_lower_bound,LFI_1984,properties,stein2016pessimistic,LOLE,stein_exponential,stein2022sensitivity}, a lower bound of the FI, as an auxiliary quantity. We then employ an auxiliary nonlinear network to transform the data in a way that its LFI approximately matches the true FI. The LFI comes with the advantage that it can be precisely estimated from much smaller statistical samples as compared to the FI.

\subsection{Linear Fisher information}
\label{subsec:Linear Fisher Information}
The LFI~\cite{TUM_LFI,properties,LOLE} is defined as 
\begin{eqnarray}
   J(\theta) & = & \left( \partial_{\theta} \langle X \rangle \right)^\top  \langle (X-\langle X \rangle) (X-\langle X \rangle)^\top \rangle\ ^{-1} \left( \partial_{\theta} \langle X \rangle \right) \nonumber\\
   & = & \left( \partial_{\theta}\bm{\mu} \right)^\top\mathbf{\Sigma}^{-1}\left( \partial_{\theta}\bm{\mu} \right)\;,
    \label{eqn:LFI}
\end{eqnarray}
where the expectation value $\langle \cdots \rangle$ is taken with respect to the likelihood $p(X;\theta)$. The vector $\bm\mu$ is the distribution mean and $\mathbf\Sigma$ its covariance matrix. Note that, for random variables in a single dimension, this inequality reduces to $J(\theta) = (\partial_{\theta}\mu)^2/\sigma^2$, with $\sigma^2$ being the variance of the random variable. For the one-dimensional case, we see that the LFI is large when a variation of the parameter $\theta$ induces a large variation of the mean $\bm\mu$, while a larger variance implies a smaller LFI. The LFI coincides with the FI, e.g., for a Gaussian distribution with respect to the mean with constant covariance \cite{parameter_estimation_literature} or for an exponential distribution with respect to the rate parameter \cite{parameter_estimation_literature,MaximumInformationStates}. What makes the LFI particularly useful is its property of being a lower bound \cite{LFI_1984,properties,stein2016pessimistic} for the FI, i.e., $J(\theta)\leq I(\theta)$.

We deal with the problem of inverting the covariance matrix in Eq.~(\ref{eqn:LFI}), which could be singular, by computing a pseudoinverse instead, based on a principal component analysis. The derivative $\partial_{\theta} \bm{\mu}$ is evaluated with a symmetric finite-difference scheme
\begin{equation}
\label{eqn:finite_difference}
    \partial_{\theta} \bm{\mu} (\theta) \approx \frac{\bm{\mu} (\theta + \Delta \theta)-\bm{\mu} (\theta - \Delta \theta)}{2 \Delta \theta}\;.
\end{equation}

In order to estimate the LFI from a finite-sized dataset, the moments $\bm\mu$ and $\mathbf\Sigma$ can be approximated by the sample mean and the sample covariance matrix, respectively. However, the resulting estimate of the LFI is biased \cite{neurons_LFI_2}, which becomes an issue, e.g., if the step size $\Delta \theta$ is very small. Given that the size of the sample is large enough to accurately estimate the covariance matrix, we compensate for the bias picked up due to a small step size by constructing an approximately unbiased estimator of the LFI (based on Ref.~\cite{neurons_LFI_2})
\begin{equation}
    \hat{J}(\theta) = \left( \partial_{\theta}\bm{\mu}_{\mathrm{S}} \right)^\top\mathbf{\Sigma}_{\mathrm{S}}^{-1}\left( \partial_{\theta}\bm{\mu}_{\mathrm{S}} \right) - \frac{2d}{N_{\mathrm{sample}} (\Delta \theta)^2}\;,
    \label{eqn:unbiased}
\end{equation}
where $\bm{\mu}_{\mathrm{S}}$ and $\mathbf{\Sigma}_{\mathrm{S}}$  are the sample mean and the sample covariance matrix, respectively, $d$ is the dimension of the random variable and $N_{\mathrm{sample}}$ is the size of the sample.

While the FI limits the precision that can be achieved with any possible estimator, the LFI yields a bound when only linear estimators are considered. It can be shown that an estimation at the best possible precision determined by the LFI is locally achieved by the so-called best linear unbiased estimator (BLUE) \cite{parameter_estimation_literature}. The term ``locally'' refers here to the fact that the BLUE is different for different values of $\theta$. Linear regression does not generally achieve the precision bound determined by the inverse LFI but performs worse. For an ANN, it is, therefore, a nontrivial task to reach the LFI when it is trained on an extended parameter domain.

\subsection{LFI maximization}
\label{subsec:LFI Maximisation}
In many cases, the LFI can be used as a good approximation of the FI. An example where the LFI coincides with the FI is the estimation of the unknown mean value from Gaussian and Poissonian distributions. This scenario occurs, e.g., in the
scattering of electromagnetic waves \cite{bouchet2021optimal,hupfl2024continuity}, where the measurements are on a fundamental
level subject to Poissonian noise. However, there are situations where the LFI is significantly smaller than the FI. A well-known example occurs when the estimated parameter is the variance of a normal distribution. When the mean and the variance are independent of each other, the mean does not at all depend on this parameter, i.e., $\partial_{\theta} \bm\mu= \mathbf{0}$ and, thus, $J=0$. In order to solve this problem and present an analysis that applies to arbitrary parameter estimation tasks, we introduce a technique that makes use of the LFI's behavior under transformations.

We introduce an algorithm that uses nonlinear transformations of the data such that the LFI of the transformed data is gradually increased and eventually converges to a value closely matching the true FI. While in previous works \cite{LFI_1984,CF_lower_bound,stein2016pessimistic,lfi_optimization_neurons,stein2016pessimistic} bounds on the FI that are tighter than the LFI itself were derived, these methods either rely on model assumptions or are computationally very demanding in high dimensions. These restrictions prevent an application of these approaches to the distributions extracted at the hidden layers of an ANN, which are, in general, complicated and high dimensional. Here, in contrast, we present a tool that allows us to construct a suitable bound numerically, using only the data at hand.

The basic idea behind our approach is that we apply a transformation to the data that encodes the same information on multiple channels. While this is very redundant and can, of course, not increase the information content, it increases the information available to the restricted class of linear estimators, which are easy to construct. With more redundancy, more information can be extracted with a linear estimator; hence, the LFI is increased. We add more and more redundancy, gradually increasing the LFI, until a convergence criterion is met.

\begin{figure}[ht!]
\centering
\includegraphics[width=0.4\textwidth]{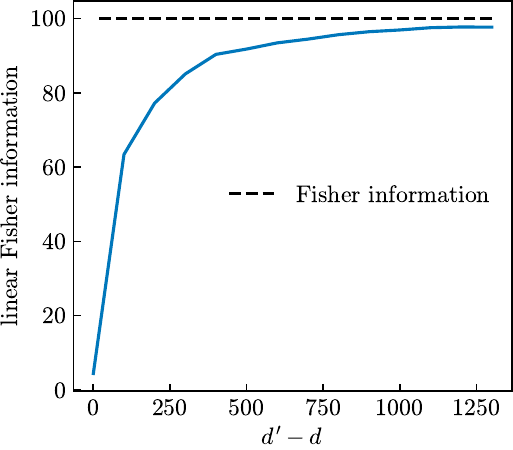}
 \caption{Maximization of the LFI for numerically generated data with vanishing initial LFI. The curve depicts the estimated LFI for consecutive embeddings. The dimension of the original dataset is $d=50$. The dimension of the transformed random variable is denoted by $d'$, which is increased in steps of $100$. The dashed line shows the true FI, which is known analytically for a normal distribution.}
 \label{fig:FIConvergence_STD}
\end{figure}

Our starting point to implement this approach is the initial random variable $X$ and the FI $I_{X}$ we want to estimate (the subscript of $I_{X}$ refers to the random variable describing the data, while the explicit dependence on the estimated parameter $\theta$ is omitted for simplicity of notation). We apply a random linear transformation $X' = \mathbf{A}X$, followed by a nonlinear, componentwise function $Y = \phi(X')+\bm{\varepsilon}$ (see Appendix \ref{section:Calculation of FI}) where $\bm\varepsilon$ is a random vector whose components are small compared to the components of $\phi(X')$. The additive term $\bm\varepsilon$ acts as a regularizer avoiding extremely narrow distributions. We then calculate the LFI of $Y$, which, for the right transformation, provides a very good approximation of the true FI $I_X$.
When the rank of $\mathbf{A}$ is equal to the dimension $d$ of $X$, $\phi$ is bijective, and moreover, when the information reduction from the added noise is small, there is no channel for a significant loss of FI in the transformation (see Appendix \ref{subsec:Sufficient Statistics and Efficiency}). For dimensions $d'$ of $X'$ greater than $d$, we observe for the distributions involved that the BLUE is in most cases able to extract more information from the transformed data compared to the original one. As a result, the LFI will likely increase. This behavior can be seen, e.g., in Fig.~\ref{fig:FIConvergence_STD}. Randomly embedding the data into a high-dimensional space and constructing a BLUE from the transformed data is analogous to an extreme learning machine \cite{rosenblatt1958perceptron,huang2015extreme} that is trained locally on a small domain of parameters $\theta$. The estimated LFI of the transformed data is, thus, an estimate of the amount of information that can be extracted with an extreme learning machine. Note that this auxiliary network cannot be used to perform the original estimation task on the whole parameter domain, as it would constitute a suitable estimator only in a small domain around a given parameter value. This is also one of the reasons why our method is computationally feasible, since constructing such a local estimator is a much simpler task than constructing a single estimator that performs well on an extended parameter domain.

Merely applying a single random embedding into a higher-dimensional space is typically not a sufficient strategy. This is because the dimension of the transformed random variable can be too small (resulting in an underestimation of the FI) or too large (resulting in an imprecise estimate of the LFI due to the finite size of the dataset). Employing a series of embeddings with a gradually increasing dimension allows us to calculate the LFI of $X'$  as a function of $d'$ and analyze the convergence of the process. We expect a monotonic curve with a plateau, i.e., a dimension, after which the increase of $J_{X'}$ is small compared to its increase in a region near $d'=d$. Since our estimator of the LFI is unbiased when sufficient data are used to estimate the LFI precisely [see Eq.~(\ref{eqn:unbiased})], our LFI curve is an unbiased estimate of the ``true'' curve, which cannot grow without bounds. Moreover, using the methods in Appendix \ref{subsec:Statistical properties of the estimated Fisher information}, we can access the statistical properties of the estimated LFI curve as a function of the dimension, the finite-difference step size, and the sample size. Specifically, Eq.~(\ref{eqn:LFI_variance}) allows us to estimate the error we make by approximating the true LFI curve with our LFI curve using a finite sample. We speak of reaching the plateau when our estimated LFI curve enters a regime of very slow growth as a function of $d'$, given that our estimated error remains small as $d'$ is varied. If the estimated error as a function of $d'$ grows too large before the plateau of the LFI curve is observed, the error can be reduced by increasing the step size or the sample size. If this is not possible, e.g., due to limitations of the experimental setup, better convergence can be achieved by measuring more data and increasing the sample size.

One scenario, where we use an alternative approach, is when the LFI is already very close to the FI. In such a case we do not observe an increase of the LFI for increasing $d'$. We then use the LFI averaged over the dimensions as the estimate for the FI. We emphasize here that, as an alternative to the extreme learning machine as the auxiliary network, one can also use other types of neural networks, with the objective of maximizing the LFI, e.g., with gradient descent. However, the loss functions that are minimized in order to identify an optimal set of weights are then typically nonconvex. This may result in the algorithm becoming trapped in a local minimum.

Our algorithm converges to the true FI, when the BLUE of the transformed data extracts all of the FI that is available in the original data, when the parameter is known to be in a small region around a parameter of reference. Satisfying this 
condition is equivalent to finding an estimator that locally achieves the CRLB, around this reference parameter. As shown in Ref.~\cite{alsing2018generalized}, such an estimator exists, even for nonexponential families of distributions. Since extreme learning machines are universal function approximators \cite{huang2015extreme}, for large dimensions of the transformed data $d'$, the auxiliary network is, in principle, able to fit this estimator and, thus, to perform at this limit. Assuming that the size of the dataset allows the LFI of the transformed data to be accurately estimated, it matches the true FI to a good approximation if $d'$ is large enough so that the corresponding extreme learning machine has a sufficient complexity to fit, e.g., the estimator presented in Ref.~\cite{alsing2018generalized}.

We demonstrate the successful application of this algorithm with the following example: We draw data from a normal distribution and employ the LFI maximization algorithm to compute the FI when the estimated parameter is the standard deviation. This is a worst-case scenario in the sense that the LFI of the initial dataset vanishes. We, thus, expect that extensive optimization is required, compared to the case where the LFI is initially close to its maximum. We observe that the distributions we encounter in our experiment are mostly of the latter type, where very little optimization is required (see Appendix \ref{section:Calculation of FI}). These concepts are illustrated in Fig. \ref{fig:FIConvergence_STD} that depicts the estimated LFI during the maximization algorithm as a function of $d'-d$. Each data point is $50$ dimensional with independent entries, drawn from a normal distribution with zero mean and a standard deviation of $\sigma = 1 \pm \Delta \sigma$. The step size is $\Delta \sigma = 0.05$ and the sample size is $100\,000$. In this simple case, the FI for the estimation of the mean $I=\frac{2N}{\sigma^2} = 100$ can be calculated analytically. The nonlinearity is described in Appendix \ref{section:Calculation of FI} and the variance of the noise term is $\sigma_{\mathrm{noise}}=0.1$. Examples of such curves for simulated and experimental data are displayed in Figs.~\ref{fig:FIConvergenceGaussian} and \ref{fig:FIConvergence}, respectively, in Appendix \ref{section:Calculation of FI}.

\section{Fisher Information Flow}
\label{Sec:Fisher Information Flow}
Equipped with this tool, we can now investigate how the FI flows through an ANN. For this purpose, we train models on two different sets of data (see Appendix \ref{sec:Neural network architectures and training parameters} for details about the datasets): The first sample we use is drawn from a numerically generated multivariate normal distribution where the parameter of interest is a location parameter, proportional to the mean. The data for the second model originate from the experiment illustrated in Fig.~\ref{fig:FI_Flow}, where the network is trained to estimate the horizontal position of the ``space invader'' from the recorded camera images. Details about the model, the data generation, the calculation of FI, and shortcomings of the LFI maximization algorithm are specified in Appendix \ref{sec:Neural network architectures and training parameters}.

\begin{figure*}[!ht]
\centering
\vspace{7mm}
\begin{overpic}[width=\linewidth]
{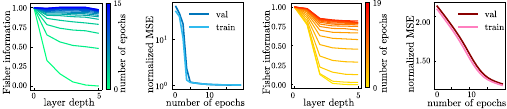}
\put(0,20){(a)}
\put(26,20){(b)}
\put(50.5,20){(c)}
\put(77.3,20){(d)}
\put(19.6,22.5){numerical data}
\put(71,22.5){experimental data}
\end{overpic}
        \caption{Fisher information calculated for a generic parameter (see Appendix \ref{sec:Neural network architectures and training parameters}): (a),(b) for numerically generated data sampled from a Gaussian distribution and (c),(d) for experimental data from the setup in Fig.~\ref{fig:FI_Flow}.
        Panels (a) and (c) show the FI for the corresponding dataset as a function of layer depth, normalized by the FI of the input dataset. With increased training of the network (see color bars for the number of training epochs), the FI in all layers approaches the FI contained in the input data. Correspondingly, the normalized learning curves drop for an increasing number of training epochs: panels (b) and (d) show the training and validation loss (the MSE cost function estimated with the training data and the validation data) as a function of the epoch number.}
    \label{fig:flow}
\end{figure*}
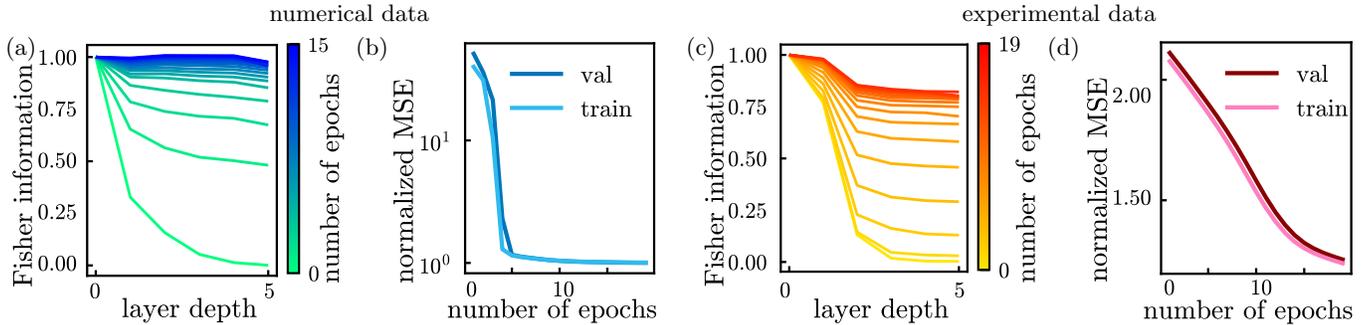

In Fig.~\ref{fig:flow}, we show the FI as a function of the layer depth at different stages of training: Figure \ref{fig:flow}(a) stands for the Gaussian data and Fig.~\ref{fig:flow}(c) for the experimental data. The corresponding mean squared error (MSE) loss functions, normalized by the FI of the dataset in the input layer (see Sec.~\ref{Fisher Information Based Early Stopping}), are shown in Fig.~\ref{fig:flow}(b) for the Gaussian data and in Fig.~\ref{fig:flow}(d) for the experimental data. The loss functions are evaluated using the training data and additionally, using an independent validation dataset that was not processed by the network during its training.

At epoch $0$, i.e., when the weights and biases are randomly initialized, a significant portion of FI is lost whenever a noninvertible transformation acts upon the data. This behavior reflects the fact that inadequately adjusted weights and biases in the ANN manifest as sinks for the flow of FI through the network. However, after sufficiently many training epochs, a large portion of the FI reaches the output layer. As the network is trained, it approaches optimal performance, where the variance of the data in the output is minimal and, as a consequence, the network conserves the FI of the data that is processed by the network.

For the Gaussian dataset, the neural network estimator achieves almost perfect conservation of FI. The network trained on the experimental imaging data achieves an output FI of around $80\%$ of the input FI after training. There is strong evidence that this failure to conserve FI is related to the range of parameters we are working with. Details on this issue and a list of factors that determine whether perfect conservation of FI can be expected are discussed in Appendix \ref{subsec:Sufficient Statistics and Efficiency}.

So far, we tracked the flow of FI evaluated at one specific parameter. However, a special feature of the FI is its locality, which means that the FI generally depends on the true parameter value. Interestingly, our method can be used to resolve the parameter dependence of the information flow itself. As we show in Appendix \ref{sec:Parameter resolved flow of Fisher information}, such a parameter-resolved analysis indeed reveals nontrivial behavior that would remain hidden with a global analysis.

With our framework based on statistical estimation theory we are able to provide a quantitative characterization of propagation of information through the neural network throughout the training process, which is a significant step toward opening the black box of ANNs when they perform estimation tasks. Given the difficulty of estimation problems with arbitrary distributions of the data, the computational cost of our method is very low and we could perform the computations on a commercial CPU. Moreover, for both the Gaussian and the experimental training datasets, the LFI itself is already close to the true FI in most of the ANNs layers, including the input layer. For these distributions the dimension in which the data are embedded does not need to be very large and the LFI maximization algorithm converges quickly.

\section{Early stopping based on Fisher information}
\label{Fisher Information Based Early Stopping}
The ultimate precision limit in the estimation process is reached, when the ANN uses all the FI $I$ contained in the input data for the estimate provided in the output layer. From this reasoning we deduce that the MSE of the estimate that the ANN can optimally provide, is limited by the CRLB:
\begin{equation}
    \mathrm{MSE} \approx  \sigma^2 \geq 1/I\;.
    \label{eqn:Early_Stopping}
\end{equation}
Assuming the ANN is capable of performing near the Cram\'er-Rao lower bound, the MSE after training should coincide with the inverse FI 
\begin{equation}
    \mathrm{MSE} \approx 1/I \;.
    \label{eqn:ES}
\end{equation}
Most importantly, this relation tells us that our ability to quantify the FI for arbitrary distributions of the input data puts us in the unique position to formulate a criterion for early stopping using only the training dataset, i.e., to stop training the network as soon as the condition in Eq.~(\ref{eqn:ES}) is met. In this way, our approach generalizes earlier studies \cite{early_stopping,CR_MAP,MMSEE,CR_informed,frequentist}, where the performance of ANNs was studied for very specific network architectures or under the assumption that the distribution of the input data is known or takes on a simple shape. Several other useful alternative strategies to conventional early stopping methods have also been developed by the deep learning community \cite{lalis2014adaptive,prechelt2002early,miseta2024surpassing,natarajan1997automated,ferro2023early,ennett2003evaluation}. What sets our method apart is that it is specifically applicable to ANNs trained to perform parameter estimation problems, and based on physical considerations.

\begin{figure}[!th]
\centering
\vspace{7mm}
\begin{overpic}[width=\linewidth]
{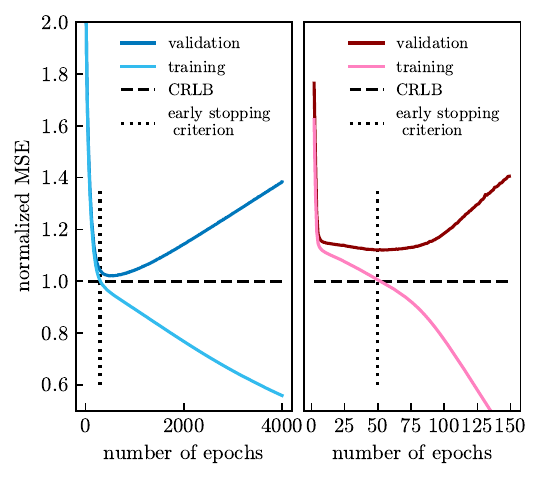}
\put(16,79){(a)}
\put(58.5,79){(b)}
\put(19.6,88.5){numerical data}
\put(63,88.5){experimental data}
\end{overpic}
        \caption{Normalized MSE as a function of number of epochs for (a) numerically generated training data with Gaussian noise and (b) data measured in the experiment. The dark blue and red lines denote the validation 
        losses, and the light blue and red lines depict the training losses. The loss function is chosen as the MSE normalized by multiplication with the FI of the input data ($\mathrm{MSE} \times I$). The dashed horizontal lines indicate where the loss is equal to $1$. The epoch number where the training loss crosses this horizontal line is marked by a vertical dashed line. This line coincides with the epoch number where the validation loss is near its minimum.}
    \label{fig:Early_Stopping}
\end{figure}
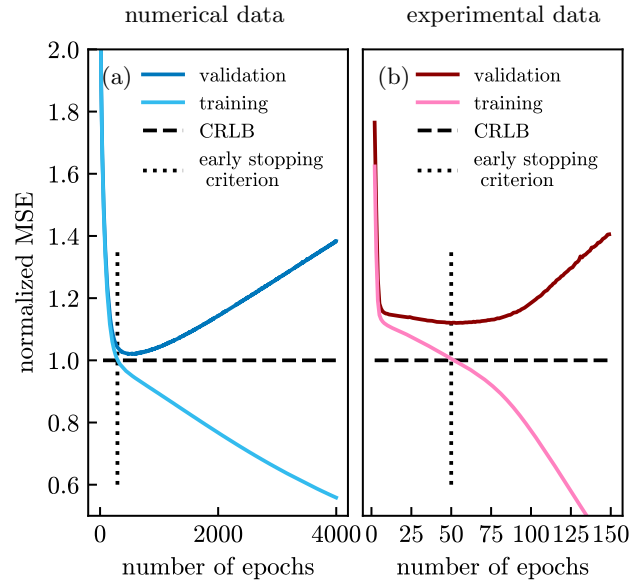

Specifically, a training loss larger or smaller than the inverse FI indicates under- or overfitting, respectively. This is illustrated in Fig.~\ref{fig:Early_Stopping}, where the MSE loss function is shown as function of the number of training epochs, for both the numerically generated training data (left) and the experimental data (right). In both cases, MSE loss functions are normalized by multiplying with the FI of the input data, so that for an estimator that achieves the CRLB the loss is equal to one.

After the training loss in Fig.~\ref{fig:Early_Stopping} crosses the horizontal line $\mathrm{MSE} \times I = 1$, the validation loss is already very close to its global minimum and further training or hyperparameter tuning cannot improve the performance of the network. We observe that with this early stopping criterion, the epoch number of minimum validation loss is estimated remarkably well directly from the training loss. For the experimental data, we observe that, although the minimum of the normalized MSE at $\mathrm{MSE} \times I = 1$ is not achieved precisely, the condition of stopping when the training loss falls below the CRLB still allows us to find a close approximation of the epoch where the validation loss is minimal. 

Note that if the complexity of the problem requires a much larger dataset than what is available in order to perform near the CRLB, this criterion is not sufficient and the model requires validation. Indeed, we need to assume that the ANN does not perform too far off the CRLB for the dataset under study. If this condition is not satisfied, $\mathrm{MSE} \times I$ can reach unity when the network is already overfitting. Details about the neural networks and the numerical methods employed in this section are provided in Appendixes \ref{sec:Neural network architectures and training parameters} and \ref{section:Calculation of FI}. Note that in Eq.~(\ref{eqn:ES}) we assume that the FI of the input data does not vary with $\theta$; otherwise, the inverse FI must be averaged over $\theta$. We also note that if a cost function other than the MSE is used, one needs to evaluate the CRLB, i.e., to compare the right-hand side of Eq.~(\ref{eqn:CRLB}) with the variance of the network's output, in addition to the cost function. Moreover, for Eq.~(\ref{eqn:ES}) to hold, the network must represent an unbiased estimator. If one expects a significant bias to occur, instead of Eq.~(\ref{eqn:Early_Stopping}), the generalized expression \cite{parameter_estimation_literature} $\sigma^2 \geq (\partial_{\theta}\langle \hat{\theta}\rangle)^2/I$ should be used as the bound for the estimator's variance. The numerator $(\partial_{\theta}\langle \hat{\theta}\rangle)^2$ in the equation above is different from $1$ if the expectation value of the estimated parameter $\langle \hat{\theta}\rangle$ does not match the true parameter, i.e., when the estimator is biased. For the networks and datasets we consider here, the assumptions of negligible bias and constant FI hold to a good approximation; thus, Eq.~(\ref{eqn:Early_Stopping}) applies.

\section{Discussion}
 \label{sec:Discussion}
With the methods introduced here, we are able to track the FI flow through an ANN. Based on this FI analysis, we observe how an ANN learns to preserve the information contained in high-dimensional image data while compressing it to a single dimension. Our approach does not require strong assumptions about the neural network architecture or the exact shape of the data distribution. A feature of our FI analysis is that it is local with respect to the parameter value. However, bounding the performance of the network, e.g., for early stopping based on training data only, requires a global measure of information. In this case, one either assumes that the FI of the training data is approximately independent of the parameter, which is the case for our datasets; alternatively, if this assumption does not hold, one needs to evaluate the FI of the training data at multiple parameters, which means that the computational cost for calculating the FI for a single parameter is multiplied with the number of parameters where the FI is evaluated. For the performance bound, the FI is averaged over the range of possible parameter values; hence, one can, instead of using a fine parameter grid, also sample random parameters, evaluate the FI at each random parameter and sum them up in order to approximate this average.

For ANNs trained to perform a parameter estimation task, our ability of estimating the FI and thereby calculating the CRLB has another simple but powerful benefit. If a model performs at or near the CRLB, we immediately know that there is no point in employing a more powerful model or searching for a better architecture, since a model that performs at this limit cannot be improved any further.

In order to illustrate that our approach is applicable to very different noise statistics, we employ in Appendix \ref{sec:Lognormal} ANNs that are trained to predict a parameter from a correlated log-normal distribution. We track the flow of FI through an ANN and demonstrate that early stopping based on the training loss works very well also for this dataset.

There is a large number of degrees of freedom for choosing the shape of the nonlinear transformations and the optimization procedure for the LFI maximization algorithm. Further research in this direction might yield more efficient techniques for estimating the FI from data. In contrast to methods for calculating the FI that entail estimating the probability density function, our approach is data driven in the sense that it profits from the size of the dataset rather than from the simplicity of the distribution. A prerequisite for our approach to work is that the dataset consists of a sufficiently large number of data points, so that the mean and the covariance matrix can be estimated accurately. Knowing the variance of our estimator of the FI (see Appendix \ref{subsec:Statistical properties of the estimated Fisher information}) and how it is affected by the choice of algorithmic parameters such as the sample size, allows us to verify whether these prerequisites are fulfilled. This insight also allows us to work out the boundaries of our algorithm and potentially also explore the regime of small datasets.

Recently, nonparametric methods \cite{non_parametric,non_parameter_ANN,edge} for estimating the FI from a finite-sized dataset that do not rely on an estimate of the probability density, have been developed. For these approaches, deriving confidence bounds is very difficult. While the accuracy of our approach cannot be predicted if the FI is unknown \cite{neurons_LFI_2}, it has the advantage that for sufficiently large datasets it always gives a pessimistic approximation of the FI \cite{stein2022sensitivity}. Such an estimate of a conservative replacement of the FI is valuable, as it not only acts as a close approximation but also yields a value for the precision that can be, at least locally (see Sec.~\ref{subsec:Linear Fisher Information}), achieved by an estimate when the data are further processed \cite{stein2022sensitivity}. Moreover, these methods typically involve functions that depend on all possible pairs of data points, i.e., $N_{\mathrm{sample}}^2$, where $N_{\mathrm{sample}}$ is the sample size, making them prohibitively expensive when a large dataset is required. The LFI maximization method involves constructing covariance matrices with a number of entries given by the squared dimension of the space in which we project the data. However, the squared dimension is in our regime of large datasets much smaller than $N_{\mathrm{sample}}^2$, which makes our LFI maximization method less expensive in terms of memory. Another drawback of these nonparametric methods is their tendency to be very sensitive with respect to the choice of the spacing between the discrete values of the parameter in the dataset, when evaluating the FI with a finite-difference scheme \cite{step_size}. We observe that our method yields inaccurate results when the spacing is very small; however, it yields close pessimistic approximations of the FI for larger spacings.

\section{Summary and outlook}
\label{sec:Summary and outlook}

Our work demonstrates how ANNs that are trained for performing estimation tasks, optimize their performance by maximizing the flow of FI from the input to the output layer. Using numerically generated data, we observe that when all of the FI stored in the input data reaches the output layer, where the estimate is produced, the optimal 
performance at the CRLB is achieved. For the case of a concrete optical experiment, involving the estimation of an object's position from noisy image data, we find that the trained network is able to channel around 80\% of FI to the output layer and, thus, comes close to the CRLB although not exactly reaching it, which we discuss in Appendix \ref{subsec:Sufficient Statistics and Efficiency}.

The nontrivial task of calculating the FI in the hidden layers is accomplished by using the quantity of LFI, which often is a close approximation of the FI and comes with the advantage of being easy to compute. In order to deal with situations where this approximation is insufficient, we present an algorithm that gradually improves upon the LFI by applying nonlinear embeddings into spaces of ever-increasing dimensionalities, until convergence is achieved. This computational tool not only allows us to study the FI flow, but also yields a criterion for early stopping, which requires only knowledge about the FI of the input data. From a conceptual point of view, our results demonstrate that, similar to the information-conserving flow of FI mediated by electromagnetic fields~\cite{hupfl2024continuity}, FI can also flow through suitably trained ANNs without any information getting lost along the way.

The opportunities for a FI flow analysis in ANNs are numerous, such as for identifying optimized network architectures, and for reducing the need for expensive trial and error approaches. Moreover, conservation of FI yields a condition for constructing an ANN layer by layer by choosing the weights such that the FI of one layer equals that of a previous layer. This could help to resolve problems such as vanishing and exploding gradients, since one does not need to propagate the gradients through the whole network. Conservation of FI can also be used as a condition to compress the data to a smaller dimension, without losing information about the parameter of interest \cite{alsing2018generalized}. As a next step, it would also be interesting to include prior information of the data about the estimated parameter. Extending our analysis to networks with skip connections such as the popular ResNet architecture \cite{he2016deep}, where the layers of the network do not show the Markov chain structure (on which we relied when assuming that the FI flow obeys the data processing inequality), would be a natural next step toward opening the black box of more complex ANN models. Moreover, we aim to apply our FI analysis to physical neural networks \cite{wright2022deep}, in order to gain a better understanding of how they learn to process information during their training \cite{momeni2024training,wanjura2024fully}.

\section{Acknowledgments}
We thank Sabine Andergassen and Thomas Gärtner for helpful discussions. L.M.R.\ was supported by the Austrian Science Fund (FWF) through Project No.\ P32300 (WAVELAND). I.S.\ and D.F.\ acknowledge financial support from the United Kingdom Engineering and Physical Sciences Research Council (EPSRC Grants No.\ EP/T00097X/1 and No. EP/Y029097/1). D.F.\ acknowledges support from the United Kingdom Royal Academy of Engineering Chairs in Emerging Technologies Scheme. The authors acknowledge the TU Wien Bibliothek for financial support through its Open Access Funding Programme.

\section{Data availability}
The experimental data \cite{weimarzenodo}
 and an exemplary code for processing the experimental data and the numerically simulated data \cite{weimargit} are publicly available.

\appendix

\setcounter{figure}{0} 
\makeatletter 
\renewcommand{\thefigure}{A\@arabic\c@figure}
\makeatother

\setcounter{table}{0} 
\makeatletter 
\renewcommand{\thetable}{A\@Roman \c@table}
\makeatother

\section{MATHEMATICAL BACKGROUND}
\label{sec:Mathematical Background}

\subsection{Sufficient statistics and conservation of Fisher information}
\label{subsec:Sufficient Statistics and Efficiency}
Sufficient statistics \cite{properties,IB_DL,FII} are transformations that preserve all the relevant features of the data with regard to the parameter of interest. 
Without much loss of generality we state that sufficiency of a statistic and preservation of FI are equivalent (in Ref.~\cite{pollard2013note}, a counterexample is constructed, which is, however, not relevant here). Finding a sufficient statistic is itself not a difficult task, as any bijective transformation preserves information. In parameter estimation, however, an additional desired property is to simultaneously compress the data to a dimension that coincides with the number of parameters to be estimated. Unless the data follow an exponential family of distributions, a nontrivial sufficient statistic that also compresses the data does generally not exist; hence, one cannot expect that it is possible to compress the data without losing FI \cite{parameter_estimation_literature}. Motivated by Refs.~\cite{IB_DL, CR_informed,frequentist,early_stopping}, we relax the conditions of sufficiency and we speak of conservation when the FI stays approximately the same as the data are transformed. Whether it is possible to approximately conserve all of the FI depends on the estimation task as discussed in the following.

FI can be lost whenever the function connecting two layers of an ANN cannot be inverted. This happens if either the dimension of the data is reduced or if the activation function \cite{python_DL}, i.e., the nonlinearity that acts upon all hidden layers' neurons, is not bijective. Training a neural network to estimate a single parameter from image data requires that the data are projected from $d$ to one dimension, where the former represents the number of pixels in the image. Hence, the network cannot always represent a sufficient statistics and one cannot expect that all of the FI present in the data is conserved as the data are processed by the neural network. Whether a model can reach the (bias-corrected) CRLB and conserve all of the FI depends on the distribution and the signal-to-noise ratio \cite{van2004detection,serbes2022threshold} but also on the range of parameters that the model is trained on \cite{papoulis1967probability,alsing2018generalized}. The broader this range is the harder it becomes for the model to reach the CRLB and to conserve the FI. One way to understand this behavior is to note that the CRLB is determined by the log-likelihood function and its first derivative with respect to the parameter. If the log-likelihood function has a complicated shape and one represents it as a Taylor series on an extended parameter interval, higher-order derivatives need to be taken into account and the CRLB can be too lose \cite{papoulis1967probability}. If the parameter range on which the neural network is trained is sufficiently small, however, these higher-order derivatives contribute only little and the CRLB becomes (approximately) reachable.

\begin{figure}[th!]
\centering
\includegraphics[width=\linewidth]{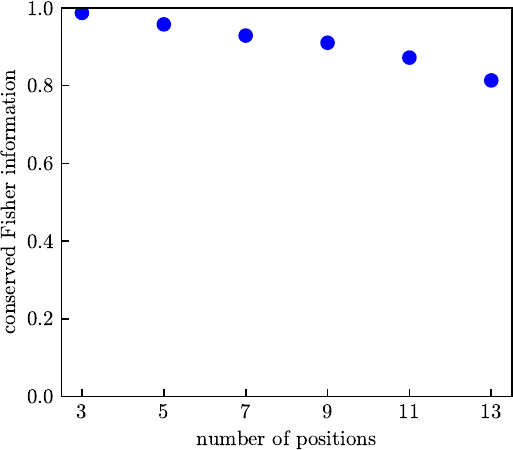}
 \caption{The fraction of the input FI that arrives at the output layer of an ANN versus the number of positions used to train the ANN. The spacing between the position is equal ($\Delta \theta = 0.3$ \textmu m); hence, more positions correspond to a broader spatial range. $13$ positions correspond to the parameter range we used to generate the results in the manuscript. We observe that the narrower the range, the more FI is conserved. The ANN
layers from input to output consist of $[432,1000,500,250,50,1]$ nodes.}
 \label{fig:FI_vs_positions}
\end{figure}

In order to investigate if the effects mentioned above play a role in the mismatch between input and output FI we observe for the experimental data, we train ANNs to estimate the position of the object, each ANN with a different number of positions shown during the training. The FI is always evaluated using only the three central positions. The fraction of the input FI that arrives at the output of the ANNs trained on different parameter ranges is shown in Fig.~\ref{fig:FI_vs_positions}.
We clearly observe that for narrower parameter ranges, more FI is conserved and when only the central three positions remain in the parameter range (those where we evaluate the FI), the mismatch disappears entirely. This result provides strong evidence that the mismatch we observed is due to the finite range of parameters we were working with.

Note that there can be other reasons for such a mismatch. Examples are a (stochastic or deterministic) error in the estimated FI, a bias picked up by the estimator that cannot by taken into account with the parameter resolution at hand, or that not enough training data are available to achieve the best performance possible.

\subsection{Statistical properties of the estimated Fisher information}
\label{subsec:Statistical properties of the estimated Fisher information}
Since in our algorithm of LFI maximization we estimate the FI using a finite-sized dataset, the estimated FI itself is a random variable. Here, we provide an analytical expression for the variance of this random variable, allowing one to gauge whether the data at hand allow for a sufficiently precise estimation of the FI. Note that our estimate of the FI is the LFI of the data after it is projected into a higher dimension (see \ref{subsec:LFI Maximisation}). The variance of our estimate is, thus, the variance of this LFI of the transformed data. In order to approximate the variance of the LFI, we carry out a derivation similar to that in Ref.~\cite{neurons_LFI_2} and find, under the assumption that the size of the sample is large 
enough to accurately estimate the covariance matrix and that the sample mean, as a sum of independent random variables, approximately follows a normal distribution, the expression
\begin{equation}
    \frac{\sigma_{\hat{J}}^2}{J^2} \approx \frac{8}{L}\left( 1 + \frac{d}{L}\right) := \eta(L)\;,
    \label{eqn:LFI_variance}
\end{equation}
where $\sigma_{\hat{J}}^2$ is the variance of the estimated LFI $\hat{J}$, $J$ the LFI, and $d$ the dimension of the random variable and we introduce $L=JN_{\mathrm{sample}}\Delta\theta^2$ with the sample size $N_{\mathrm{sample}}$. To estimate the FI, we gradually increase $d$ until convergence is achieved. The results from the algorithm are precise if the fluctuations of the estimated LFI are small compared to the LFI, i.e., $\eta(L) \ll 1$ for the largest value of $d$ we chose in the algorithm. Note that if the LFI curve as a function of the dimension is not sufficiently smooth, we average over a few dimensions to smoothen the curve and identify the plateau more easily; it is reasonable to assume that this summation of LFIs does not increase the standard deviation.

From Eq.~(\ref{eqn:LFI_variance}), we learn that the quantity $L$ should be large in order to precisely estimate the FI. This means that the sample size must be large enough to compensate for the squared finite-difference step size, which must be chosen small enough such that the discretization error remains small. 

We also note that the term in Eq.~(\ref{eqn:LFI_variance}) that is proportional to $d$ decreases with $1/L^2$, while the term independent of $d$ only decreases with $1/L$. Hence, for large $L$, the uncertainty of the estimated FI depends only weakly on the dimension of the projection. 

\section{EXPERIMENTAL METHODS}
\label{sec:experimental_methods}

\subsection{Optical setup}

Broadband light is generated by a fiber optic illuminator (Oriel, Model No. 77501) coupled to a multimode fiber. Light is outcoupled and attenuated using a neural density filter (optical density 4) before being used to illuminate the sample. On this sample, a drawing of a space invader was engraved onto a chromium layer (CAD/Art Services). The position of the sample is controlled by a motorized stage (PI M-122.2DD1). Transmitted light is collected by a microscope objective (Olympus Plan N $\times$20, 0.4 NA) and imaged using a tube lens ($f=150$\,mm) onto a camera (PCO Panda 4.2 sCMOS).

\subsection{Acquisition procedure}
\label{subsec:Acquisition procedure}

The sample is linearly translated along the x direction with a step size of $0.3$\,\textmu m and images are collected for 13 positions (measurements are taken over a broader range of positions but the outer positions are discarded to keep the data set at a moderate size). The range of positions covered by the object is, thus, comprised between $x=-1.8$\,\textmu m and $x=+1.8$\,\textmu m. The image dimension is set to $32\times64$ pixels, with a binning procedure such that each pixel corresponds to the sum of $4\times4$ physical pixels of the camera. The exposure time is set to $10$\,\textmu s, which results in a very low signal-to-noise ratio. $300 \,000$ images are measured for the three central positions ($x=-0.3$\,\textmu m, $x=0$\,\textmu m, and $x=+0.3$\,\textmu m), and $50 \, 000$ images are measured for all other positions. Finally, a set of images is measured for different sample positions using a long exposure time ($100$\,ms) for illustrative purposes. 

\subsection{ Noise of the experimental data}
\label{subsec:Sources of noise}
\begin{figure}[!th]
\centering
\vspace{7mm}
\begin{overpic}[width=\linewidth]
{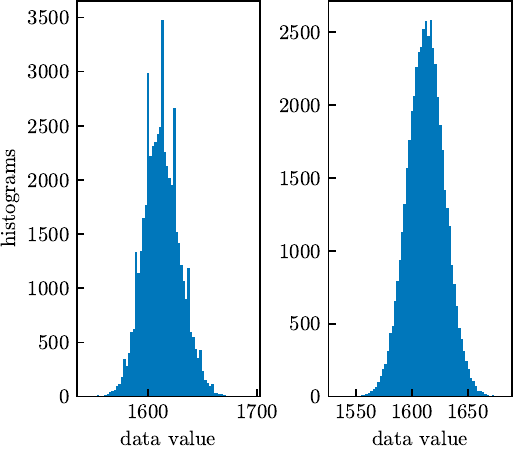}
\put(6,89.5){(a)}
\put(55,89.5){(b)}
\end{overpic}
        \caption{(a) shows histograms for a generic marginal distribution of the experimental data (a pixel at the center of the image). The number of histogram bins is $70$. We use a total of $50\ 000$ data points. (b) shows histograms for a Gaussian distribution with the same mean and variance as the experimental marginal distribution, with the same number of data points and number of bins.}
    \label{fig:camera_marginal}
\end{figure}

In Fig.~\ref{fig:camera_marginal} we visualize the distribution of our experimental data by showing a histogram fit to the distribution of a generic camera pixel. We compare this with a sample of the same size from a Gaussian distribution with the same mean and variance, in order to visualize the difference between the two distributions.
While the shape of the distribution roughly follows a Gaussian envelope, this distribution actually significantly deviates from a Gaussian distribution and we cannot easily come up with an accurate model for it.

In order to determine what is the dominant source of noise in our experiment, we measured the standard deviation of the camera noise $\sigma_{\mathrm{cam}}$, in the absence of incoming light, as a function of the exposure time. We observed that the standard deviation is constant ($\sigma_{\mathrm{cam}}\simeq 15\,\mathrm{DN}$) for exposure times lower than $10$\,ms. This indicates that the contribution of readout noise is much larger than that of thermal noise in our experiment, for which the exposure time was set to $10$\,\textmu s. In addition, we can estimate the expected shot noise contribution. In our experiment, the level of signal is approximately $s \simeq 2\,\mathrm{DN}$ for each bright pixel. Taking into account the measured analog-to-digital factor for our camera $\eta_{\mathrm{AD}}=0.55$\,e$^{-}$/DN (photoelectrons per digital unit), the expected standard deviation due to shot noise is $\sqrt{s/\eta_{\mathrm{AD}}}\simeq2$\,DN. This value being significantly lower than $\sigma_{\mathrm{cam}}$, we conclude that the contribution of readout noise is larger than that of shot noise in our experiment.

\section{DATA GENERATION}
\label{sec:Data generation}
For each ANN we list the details about the generation of the training and testing datasets.

\subsection{Generation of the Gaussian dataset for observing the Fisher information flow}
\label{subsec:Generation of the Gaussian data set for observing the Fisher information flow}
We sample $d=16\times 16 = 256$-dimensional data from a multivariate normal distribution with a diagonal covariance matrix $\mathbf\Sigma = \mathbb{I}_{d\times d}$ and a mean vector proportional to the parameter of interest, $\bm\mu(\theta) = \theta \times (1,1,...)^\top$.
We choose $31$ distinct $\theta$ values on an equally spaced grid, ranging from $\theta = -1$ to $\theta=1$, and sample $20\,000$ data points for each parameter value.

80\% of the dataset is used to train the ANN, while the remainder is reserved for validation. A separate dataset is prepared for evaluating the FI flow. To approximate the derivative in the expression for the FI by a symmetric finite-difference scheme, we first generate data on an equally spaced grid, but with three grid points $\theta \in \{-0.03, 0, 0.03\}$. 150\ 000 data points are generated per grid point.

\subsection{Generation of the experimental dataset for observing the Fisher information flow}
\label{subsec:Generation of the experimental data set for observing the Fisher information flow}
As described in Appendix \ref{subsec:Acquisition procedure}, the training dataset consists of $50\,000$ images for each of the $13$ positions of the scatterer. Observing that information about the scatterer arrives only at a central region of the camera, we crop the images symmetrically around the center, such that the size of the images is reduced from $32\times 64$ to $18\times 24$. This procedure has a small effect on the information content of the data but reduces the memory required to store the datasets. Before it is fed to the network, the data are standardized by subtracting the intensity averaged over all images and dividing by the corresponding standard deviation. This invertible transformation does not alter the information content. 80\% of the dataset is used to train the ANN while the remainder is reserved for validation.

The testing data used to track the FI flow are measured and processed in the same way but consist of 150\,000 images per position for a total of three equally spaced positions at $\theta \in \{\SI{-0.3}{\micro\metre}, \SI{0}{\micro\metre}, \SI{0.3}{\micro\metre}\}$. 

\subsection{Generation of the Gaussian dataset for observing Fisher-information-based early stopping}
\label{subsec:Generation of the Gaussian data set for observing Fisher information based early stopping}
The dataset is equivalent to that in Appendix \ref{subsec:Generation of the Gaussian data set for observing the Fisher information flow} but with $101$ parameter grid points and with $7000$ data points for each grid point and ranging from $\theta=-1.5$ to $\theta=1.5$.

\subsection{Generation of the experimental dataset for observing Fisher-information-based early stopping}
\label{subsec:Generation of the experimental data set for observing Fisher information based early stopping}
Compared to the dataset from Appendix \ref{subsec:Generation of the experimental data set for observing the Fisher information flow}, we use only $11\,000$ images per position.

 \section{NEURAL NETWORK ARCHITECTURES AND TRAINING PARAMETERS}
\label{sec:Neural network architectures and training parameters}

The details about the ANNs are listed in Table \ref{table:1}.

\begin{table}[h!]
\begin{center}
\begin{tabular}{ | m{4.3em} | m{1.7cm}| m{1.5cm} | m{1.4cm} | m{1.5cm} | } 
  \hline
    & Gaussian FI flow & Exp. FI flow & Gaussian ES & Exp. ES \\ 
  \hline
  Nodes & $256 - 150 -100- 50 - 25 - 1 $ & $432 - 800 -100- 25 - 5 - 1 $ &  $256 - 1000 -500- 300 - 50 - 1 $ & $432 - 5000 -1500- 250 - 150 - 1 $\\ 
  \hline
  Loss & MSE & MSE & MSE & MSE \\ 
  \hline
  Optimizer & ADAM & ADAM & ADAM & ADAM \\ 
  \hline
  Batch size & 128 & 128 & 128 & 128 \\ 
  \hline
 Learning rate & $ 1\times 10^{-6}$ & $5\times 10^{-7}$ & $1\times 10^{-7}$ & $1\times 10^{-7}$ \\ 
  \hline
\end{tabular}
\end{center}
\caption{Neural network architectures and training parameters for the ANNs employed to observe the flow of FI and early stopping based on FI, for both the Gaussian training data and the experimental training data.}
\label{table:1}
\end{table}

 \section{CALCULATION OF FISHER INFORMATION}
\label{section:Calculation of FI}
The LFI itself already yields satisfying results in most of the ANN's layers. We employ the LFI maximization algorithm to find corrections to the LFI. Especially in early stages of the training there is no guarantee that the LFI is a good approximation to the true FI; thus, the LFI maximization algorithm yields more accurate and reliable results compared to using only the LFI. The linear transformations for embedding into a higher-dimensional space are random matrices with independent and identically distributed entries following a normal distribution with 0 mean and a standard deviation of $\sigma=1$. We then apply the nonlinearity
\begin{equation}
    \phi(x) = \begin{cases} 
      \alpha \times x & x\leq 0 \\
        x & x > 0 \;,
   \end{cases}
\end{equation}
with $\alpha = 0.7$ to each component of the transformed random variable and eventually add noise following a normal distribution with $0$ mean and a standard deviation of $\sigma_{\mathrm{noise}}$. In each iteration, we increase the dimensionality by $\Delta d$. As the criterion for convergence, we use that the difference between the smallest and the largest value of the LFI within the previous three iterations must be smaller than that of the first three iterations by a factor of $\kappa$. If during the first three iterations the increase of the LFI is smaller than a fraction of $\gamma$ of the LFI with $d'=d$, we stop the maximization procedure and choose the LFI averaged over the first three estimates as the estimate of the true FI. The convergence criteria could not be fulfilled for all layers and epochs of the networks trained on the different datasets; thus, the LFI itself is in these cases used as a rough approximation of the true FI. If the optimized LFI is below the LFI of the dataset, we use the LFI as the estimate for the FI. This case occurs whenever the LFI is very close to the FI and the algorithm slightly underestimates the FI due to the added noise. In Table \ref{table:2} we list the parameters of the LFI maximization algorithm together with the exceptions, where the algorithm did not converge.

\begin{table}[th!]
\begin{center}
\begin{tabular}{ | m{5.8em} | m{1.7cm}| m{1.cm} |  } 
  \hline
    & Gaussian FI flow & Exp. FI flow  \\ 
  \hline
  $\Delta d$ & 30 & 30 \\ 
  \hline
  $\kappa$ & 0.1 & 0.1 \\ 
  \hline
  $\gamma$ & 0.05 & 0.05 \\ 
  \hline
  $\sigma_{\mathrm{noise}}$ & 0.01 & 0.01 \\ 
  \hline
  Exceptions & Output layer at epoch 0 & Layer 2: epoch 0 and 1; layer 4: epoch 0 \\ 
  \hline
\end{tabular}
\end{center}
\caption{Parameters used for calculating the FI with the LFI maximization algorithm. Data are extracted at each layer of the ANNs that are trained to observe the FI flow.}
\label{table:2}
\end{table}

Note that it is not necessary to apply the LFI-based algorithm to the input layer of the ANN trained with the Gaussian training dataset as the FI of this dataset is known analytically (see Sec.~\ref{subsec:Linear Fisher Information}).

\begin{figure}[ht!]
\centering
\includegraphics[width=0.4\textwidth]{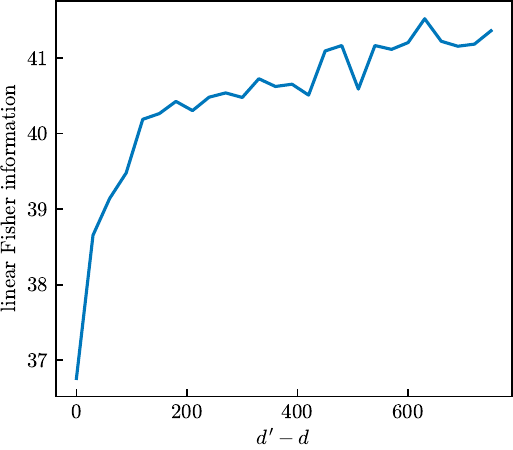}
 \caption{LFI during the LFI maximization for Gaussian data. The data are extracted at the second hidden layer before the first epoch of training. The curve depicts the estimated LFI for consecutive embeddings. The dimension of the original dataset is $d=100$, and it is increased in steps of 30.}
 \label{fig:FIConvergenceGaussian}
\end{figure}

\begin{figure}
\centering
\includegraphics[width=0.4\textwidth]{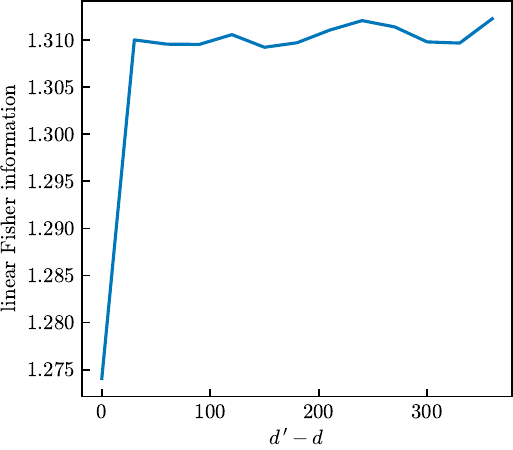}
 \caption{LFI during the LFI maximization for the experimental training data set. The data are extracted at the output layer as the network is fully trained. The curve depicts the estimated LFI for consecutive embeddings. The dimension of the original dataset is $d=1$, and it is increased in steps of 30.}
 \label{fig:FIConvergence}
\end{figure}

Figures \ref{fig:FIConvergenceGaussian} and \ref{fig:FIConvergence} show the LFI as a function of $d'-d$ for an exemplary layer of the network trained on the Gaussian dataset and one of the network trained on the experimental dataset. Note that the exact shapes of these curves depend also on the magnitude of the noise that is added for regularization.

\section{PARAMETER-RESOLVED FLOW OF FISHER INFORMATION}
\label{sec:Parameter resolved flow of Fisher information} 
\begin{figure}
\centering
\includegraphics[width=0.4\textwidth]{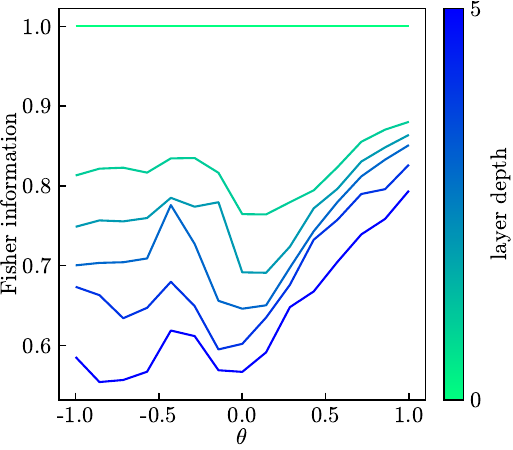}
 \caption{Fisher information flow through an ANN trained on the Gaussian dataset. The curves show the (normalized) Fisher information as a function of the estimated parameter $\theta$. Each curve corresponds to a certain layer of the ANN, which is indicated by the color of the curve. The network architecture is the same as that for the flow of FI for a single parameter. Note that we stop the training at an early epoch such that it is possible to distinguish the different curves. The FI is, thus, not conserved in this example.}
 \label{fig:parameter_resolved_flow}
\end{figure}

As the FI is local in the parameter-i.e., it can be different for different values of $\theta$ we can access the local flow of information through an ANN. To demonstrate this, we train an ANN on the Gaussian dataset where the estimated parameter is proportional to the mean and calculate the layerwise FI as a function of this parameter. The neural network architecture and the specifics of the training dataset are the same as those we used to observe the FI flow. We evaluate the FI at $15$ points in the range $\theta \in \{-1,1\}$ and generate for each point three samples of size $100\,000$, one at the point of observation $\theta$, one at $\theta-0.1$ and one at $\theta+0.1$. The result is shown in Fig.~\ref{fig:parameter_resolved_flow}.
We observe that the flow of FI is a highly nonuniform and nonsymmetric function of the parameter. The ANN, thus, learns to process data at different parameters differently. The parameters we use for our algorithm for calculating the FI are the same as in Table \ref{table:2}, except for $\sigma_{\mathrm{noise}}=0.05$.

\section{TESTING OUR METHODS ON DATA FOLLOWING A CORRELATED LOG-NORMAL DISTRIBUTION}
\label{sec:Lognormal}
We train here ANNs on a dataset following a correlated log-normal distribution. Log-normal distributions apply to scenarios very different from our imaging experiment, such as earth and environmental science \cite{andersson2021mechanisms}. The log-normal distribution is skewed and has a right tail that decays slowly, thus allowing for “outliers”, resulting in
a harder parameter estimation task.

\begin{figure}
\centering
\includegraphics[width=0.4\textwidth]{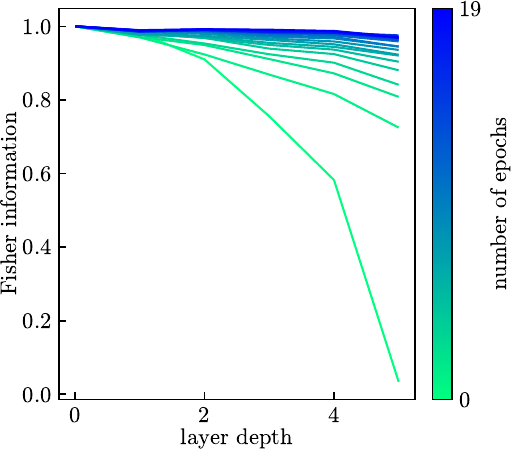}
 \caption{Fisher information calculated for a generic parameter ($\theta=0$) for data sampled from a correlated log-normal distribution. The curves show the FI as a function of layer depth, normalized by the FI of the input dataset. With increased training of the network (see the color bar for the number of training epochs), the FI in all layers
approaches the FI of the input data. The ANN’s layers from input to output consist of [10, 150, 100, 50, 25, 1] nodes.}
 \label{fig:FIFLowLognormal}
\end{figure}

\begin{figure}
\centering
\includegraphics[width=0.4\textwidth]{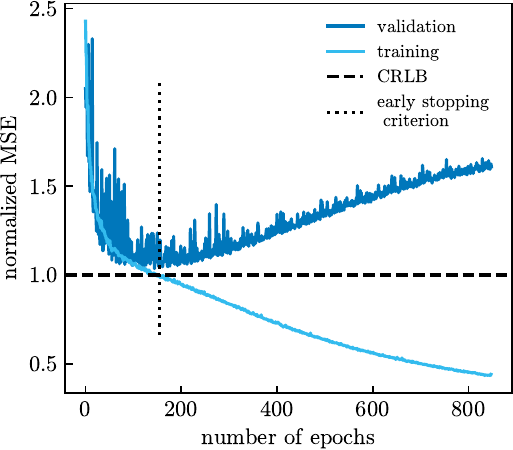}
 \caption{Normalized MSE as a function of the number of epochs for the data following a correlated log-normal distribution. The dark blue line denotes the validation loss and the light blue line the training loss. The
loss function is chosen as the MSE normalized by multiplication with the FI of the input data. The dashed horizontal line indicates where the loss is equal to 1. The epoch number where the training loss crosses this horizontal line is marked by a vertical dashed line. This
line coincides with the epoch number where the validation loss is near its minimum. The ANN’s layers from input to output consist of [10, 1500, 1000, 350, 150, 1] nodes.}
 \label{fig:ES_Lognormal}
\end{figure}

Our procedure for generating the data is the
following: We sample Gaussian data analogously to Appendix~\ref{sec:Data generation}, with a dimension of $d=10$. We now introduce correlations between the components by multiplying each data point with a Gaussian random matrix (with independent and identically distributed entries of $0$ mean and a standard deviation of $0.1$). Afterward, we exponentiate each component of the data. The marginal distribution of each component then follows a log-normal distribution with the components being dependent of each other. While the data follow now a complicated distribution, we still have access to the exact FI, which is equal to that of the underlying Gaussian distribution, since the transformation from the Gaussian random variable to the log-normal random variable is bijective and, thus, preserves the FI. We train an ANN to predict the location parameter of the underlying Gaussian (see Appendix~\ref{sec:Data generation}) and show the flow of FI in Fig.~\ref{fig:FIFLowLognormal}.
We observe that, after a few epochs of training, the ANN manages to conserve the FI to a good approximation. In Fig.~\ref{fig:ES_Lognormal}, we show the learning curves, i.e., the MSE estimated using our training and validation dataset, for an ANN trained on the log-normal dataset.
The model is much larger compared to that we used to calculate the flow of FI, in order to detect overfitting more easily. We observe that our early stopping criterion based on the training loss allows for a very good estimate of the epoch after which the model starts overfitting. The parameters we used for the neural networks and to calculate the FI are listed in Tables \ref{table:3} and \ref{table:4}. For observing the FI flow and also for our early stopping criterion we generate training data consisting of $20\ 000$ data points at $21$ equally spaced parameters in the range of $\theta \in\{ -7,7\}$. For calculating the FI we use $150\,000$ data points at three positions with a spacing of $\Delta\theta=0.1$.
\begin{table}[th!]
\begin{center}
\begin{tabular}{ | m{5.8em} | m{1.7cm} |  } 
  \hline
     & FI flow  \\ 
  \hline
  $\Delta d$ & 50  \\ 
  \hline
  $\kappa$ & 0.1 \\ 
  \hline
  $\gamma$ & 0.05  \\ 
  \hline
  $\sigma_{\mathrm{noise}}$ & 0.01  \\ 
  \hline
  Exceptions & Output layer at epoch 0 epoch 0 \\ 
  \hline
\end{tabular}
\end{center}
\caption{Parameters used for calculating the FI with the LFI maximization algorithm for the log-normal data. Data are extracted at each layer of the ANNs that are trained to observe the FI flow.}
\label{table:3}
\end{table}
\begin{table}[h!]
\begin{center}
\begin{tabular}{ | m{4.3em} | m{2.9cm}|m{3.6cm}|} 
  \hline
    & FI flow & ES \\ 
  \hline
  Nodes & $10, 150, 100, 50,
25, 1$ & $10, 1500, 1000, 350, 150, 1$ \\ 
  \hline
  Loss & MSE & MSE  \\ 
  \hline
  Optimizer & ADAM & ADAM\\ 
  \hline
  Batch size & 128 & 128 \\ 
  \hline
 Learning rate & $1\times 10^{-4}$ & $5\times 10^{-6}$  \\ 
  \hline
\end{tabular}
\end{center}
\caption{Neural network architectures and training parameters for the ANN employed to observe the flow of FI and early stopping based on FI for the log-normal data}
\label{table:4}
\end{table}

\newpage
\bibliography{references}

\end{document}